\documentclass[conference,a4paper]{IEEEtran}
\IEEEoverridecommandlockouts
\usepackage{cite}
\usepackage{amsmath,amssymb,amsfonts}
\usepackage{algorithmic}
\usepackage{graphicx}
\usepackage{textcomp}
\usepackage{xcolor}
\def\BibTeX{{\rm B\kern-.05em{\sc i\kern-.025em b}\kern-.08em
    T\kern-.1667em\lower.7ex\hbox{E}\kern-.125emX}}
\begin{document}

\title{Optimized Hybrid Focal Margin Loss for Crack Segmentation
}

\author{\IEEEauthorblockN{1\textsuperscript{st} Jiajie Chen}
\IEEEauthorblockA{
\textit{MegaRobo Technologies Co., Ltd.}\\
Shanghai, China \\
cjj3779@hotmail.com}
}

\maketitle

\begin{abstract}
Many loss functions have been derived from cross-entropy loss functions such as large-margin softmax loss and focal loss. The large-margin softmax loss makes the classification more rigorous and prevents overfitting. The focal loss alleviates class imbalance in object detection by down-weighting the loss of well-classified examples. Recent research has shown that these two loss functions derived from cross entropy have valuable applications in the field of image segmentation. However, to the best of our knowledge, there is no unified formulation that combines these two loss functions so that they can not only be transformed mutually, but can also be used to simultaneously address class imbalance and overfitting. To this end, we subdivide the entropy-based loss into the regularizer-based entropy loss and the focal-based entropy loss, and propose a novel optimized hybrid focal loss to handle extreme class imbalance and prevent overfitting for crack segmentation. We have evaluated our proposal in comparison with three crack segmentation datasets (DeepCrack-DB, CRACK500 and our private PanelCrack dataset). Our experiments demonstrate that the focal margin component can significantly increase the IoU of cracks by 0.43 on DeepCrack-DB and 0.44 on our PanelCrack dataset, respectively.
\end{abstract}


\section{Introduction}
Cracks are common surface defects that occur everywhere. Minor cracks on the surface of an object can be easily mixed up with complex backgrounds. But even tiny cracks can pose immeasurable security risks to space equipment, sophisticated electronics and more. Returning defective goods to the factory also increases post-sale costs. Since the crack data satisfies the characteristics of extreme class imbalance, crack detection application is chosen as experimental validation.
 
 Deep neural networks have yielded remarkable results in various computer vision tasks such as classification and object detection. While these methods can detect whether an image contains cracks or not, in general, image segmentation methods are more elaborate. Numerous works focus on designing better networks with encoder-decoder architectures, e.g., U-Net, SegNet \cite{SegNet}, V-Net \cite{V-net} and UNet++ \cite{Unet++}. Several works have improved the performance of these networks using the attention mechanism, e.g., Attention U-Net \cite{Att-Unet} and Attention U-Net++ \cite{Att-Unet++}. Some other works focus on making the encoder stronger, e.g., E-Net \cite{E-net} adopts early downsampling, which heavily reduces the input size in the first two blocks for real-time purposes. Eff-UNet \cite{Eff-UNet} uses EfficientNet \cite{Efficientnet} as an encoder in combination with a U-Net decoder. EfficientUNet++ \cite{Eff-Unet++} is based on EfficientNet and U-Net++. TransUNet \cite{Transunet} applied vision transformer (ViT) \cite{ViT} as encoder. 
 Recently, the proposal of decoder-part with residual blocks \cite{OptimizedCrack} has shown new state-of-the-art results on segmentation of road surface cracks and made it possible to accurately segment cracks in industrial field.
 
 In image segmentation, the design of the loss function is as important as the network design. Various loss functions have been proposed to address the class imbalance issue. One approach is to improve the cross-entropy loss, e.g., weighted cross-entropy loss \cite{WCE}, focal loss \cite{Focalloss}, asymmetric focal loss \cite{Overfitting}. Since the use of dice loss in \cite{V-net} for image segmentation, many works have turned to improve it, e.g., Tversky Index \cite{TverskyLoss}, Focal Tversky \cite{Focaltversky}, Log-Cosh dice\cite{Log-Cosh-Dice}. More recent works combine cross-entropy-based loss and dice-based loss as a compound loss, such as combo loss \cite{ComboLoss}, dice focal loss \cite{AnatomyNet}, hybrid focal loss \cite{Focus-Unet} and unified focal loss \cite{Unified-Focal}. Li \cite{Overfitting} suggested that class imbalance in the data leads to overfitting, and the regularization method is another effective way to deal with overfitting and class imbalance, which is different from the focal method. However, none of these recent works apply regularizers to entropy-based component loss. 
 
 To overcome aforementioned issue, entropy-based loss is split into regularizer-based entropy loss and focal-based entropy loss. Inspired by Unified Focal  \cite{Unified-Focal}, we propose Focal Margin to optimize the entropy-based component loss in these works. Our experiment data demonstrate the proposed loss function can significantly improve the performance on crack segmentation.

\section{Related Work}
\subsection{Networks for Crack Segmentation}
Zou $et\ al\ldotp$\cite{DeepCrack-Zou} proposed a SegNet-based DeepCrack in which the decoders employ unpooling to upsample the features. Another version of DeepCrack architecture proposed by Y.Liu \cite{DeepCrack-Liu} uses a VGG backbone and concatenates all the side output applied by deep supervision \cite{DeepSupervision}. In recent work, a re-designed decoder was proposed that can be added to various backbones such as VGG \cite{VGG}, ResNet \cite{ResNet} and EfficientNet \cite{OptimizedCrack}. Nearest neighbor up-sampling is used to increase the spatial size, which can then be concatenated with the output of the encoder at a particular level before being fed into the decoder block. Each decoder block except level one contains a standard Conv-BN-ReLU sequence followed by two residual blocks \cite{OptimizedCrack}. Two 3x3 Conv-BN-ReLU sequences are applied at level one to still extract features. Evidence in \cite{OptimizedCrack} shows that the application of residual blocks and nearest neighbor up-sampling in the decoder blocks can significantly improve the crack segmentation results.

\subsection{Losses for Class Imbalance}
A commonly used loss function is the BceDice ($L_{bcedice}$), a loss function with a combined binary cross-entropy (Bce) and dice coefficient (Dice) \cite{Sorenson-Dice}. In binary class segmentation, the $L_{bcedice}$ used in \cite{OptimizedCrack} is given as follows:
\begin{equation}
    L_{bcedice}(P,T) = -\frac{1}{N} \sum_{c=1}^{}(T_c\cdot \log_{}{P_c} + \frac{2\cdot T_c\cdot P_c + 1}{T_c+P_c + 1} )
\end{equation}
For the  class c=1, $P_c$ represents the model prediction while $T_c$ represents the ground truth for that class. The smoothing term 1 in the dice loss is added to ensure that negative samples also contribute to the training.

Li proposed that class imbalance in the data can lead to overfitting of the rare foreground class \cite{Overfitting}. Their work shows that the distribution of the activated logits of rare class shift towards and even across the decision boundary, resulting in a loss of sensitivity \cite{Overfitting}. To this end, asymmetric modifications on losses and training strategies were applied.

Asymmetric focal loss is one of the proposed modifications in \cite{Overfitting} to address the observed overfitting of neural networks under class imbalance. Since the foreground class is rare, the entropy loss contributed by the foreground is already sufficiently small compared to the large background entropy loss. It is helpful to remove the loss attenuation for the foreground class from the focal loss \cite{Focalloss}, which leads to the following asymmetric focal loss:
\begin{equation}
\mathop{L_{aF}}\limits_{c=1}(P,T)  = - \frac{1}{N}T_{c}\cdot{\log_{}{P_{c}} } - \frac{1}{N}P_{c}^{\gamma }\cdot (1-T_{c})\cdot{\log_{}{(1-P_{c})} }\label{eq.2}
\end{equation}

Another asymmetric loss analyzed in \cite{Overfitting} was the modification of the large margin loss \cite{LargeMargin}. Considering that unseen foreground class may shift toward the background class, a margin is set for rare foreground class as a regularizer to mitigate the bias of class imbalance, which leads to the following asymmetric large margin loss:
\begin{equation}
        \mathop{L_{aM}}\limits_{c=1}(P,T)  = - \frac{1}{N}T_{c}\cdot\log_{}{\hat{P_c}}  - \frac{1}{N} (1-T_{c})\cdot{\log_{}{(1-P_{c})} }\label{eq.3}
\end{equation}

Salehi further split the denominator of Dice coefficient \cite{TverskyLoss} into $T_c\cdot P_c + (1-P_c)\cdot T_c$ + $P_c\cdot (1-T_c)$, in which $(1-P_c)\cdot T_c$ represents false-negatives (FNs) and $P_c\cdot (1-T_c)$ represents false-positives (FPs). By adding coefficients $\alpha$ and $\beta$ to FNs and FPs, Tversky Index denotes as follows:
\begin{equation}
    TI(P,T)=\sum_{c=1}^{}\frac{T_c\cdot P_c + \gamma}{T_c\cdot P_c + \alpha(1-P_c)\cdot T_c+\beta P_c\cdot (1-T_c)+\gamma}
\end{equation}
Especially, when $\alpha$ = $\beta$ = 0.5, the Tversky Index becomes the Dice coefficient. Tverysky Index can be adapted to handle imbalanced data by adjusting $\alpha$ and $\beta$ to selectively focus on FNs or FPs. Although the weights of FNs and FPs can be adjusted in Tversky Index, it is still a linear loss function.

Milletari proposed a novel dice loss \cite{V-net} that simply squared the $P_c$ and $T_c$ in the denominator and turned dice from a linear function to a nonlinear one. Assuming that $P_c$ is highly close to 0, the squared $P_c$ will be even closer to 0. When $P_c$ is close to 1, the squared $P_c$ does not decrease too much. The power operation makes the loss more focused on hard samples:
\begin{equation}
    DL(P,T)=1-\frac{1}{N} \sum_{c=1}^{}\frac{2\cdot T_c\cdot P_c + \gamma}{T_c^2+ P_c^2 +\gamma}
\end{equation}

Another nonlinear variant of the Tversky loss is the Focal Tversky \cite{Focaltversky}, where the TI is first performed and then the exponential $\gamma$ is added directly on top of the 1-TI:
\begin{equation}
    \mathop{L_{FT}}\limits_{c=1}=\sum_{c}(1-TI_c)^{\gamma}
\end{equation}

Yeung \cite{Unified-Focal} recently summarized the derivation of dice-based and cross-entropy-based losses and proposed a unified focal loss to handle class imbalanced medical segmentation. Prior to unified focal loss, they proposed hybrid focal loss \cite{Focus-Unet}, a combination of focal loss ($L_F$) and focal tversky loss ($L_{FT}$):
\begin{equation}
      L_{HF}=\lambda L_F + (1-\lambda)L_{FT}
\end{equation}
They then mimicked the idea of asymmetric modification in \cite{Overfitting} and define the modified asymmetric focal tversky loss as follows:
\begin{equation}
    L_{aFT} = \sum_{c\ne r }^{} (1-TI) + \sum_{c=r}(1-TI)^{1-r}
\end{equation}
Finally, the parameters $\alpha$ in the focal loss and $\alpha$ and $\beta$ in the focal tversky loss are unified that using a single $\delta$, since these parameters are all for class weighting. The attenuation parameter $\gamma$ in focal loss and the enhancement parameter $\gamma$ in the focal tversky loss are also unified. The unified focal loss ($L_{sUF}$) and its corresponding asymmetric modifications ($L_{aUF}$) are described as follows:
\begin{equation}
    L_{sUF} = \lambda L_{F} + (1-\lambda)L_{FT}
\end{equation}
\begin{equation}
    L_{aUF} = \lambda L_{aF} + (1-\lambda)L_{aFT}
\end{equation}

\section{Methodology}
The derivation of dice-based and cross-entropy-based loss deeply depends on the degree of class imbalance. We can infer that further improvements to the derivation can be applied when the ratio is 1:20 or lower. Although asymmetric modifications have been proposed in \cite{Overfitting}, detailed relations between asymmetric loss functions have not yet been presented.
Inspired by the unified focal loss \cite{Unified-Focal}, we propose the asymmetric focal margin loss ($L _{aFM}$) which establishes the connection between the asymmetric focal loss and the asymmetric large margin loss to unify the two loss functions. 
\begin{figure}[!htb]
\setlength{\belowcaptionskip}{0.1cm}   
\centering 
\includegraphics[width=0.45\textwidth]{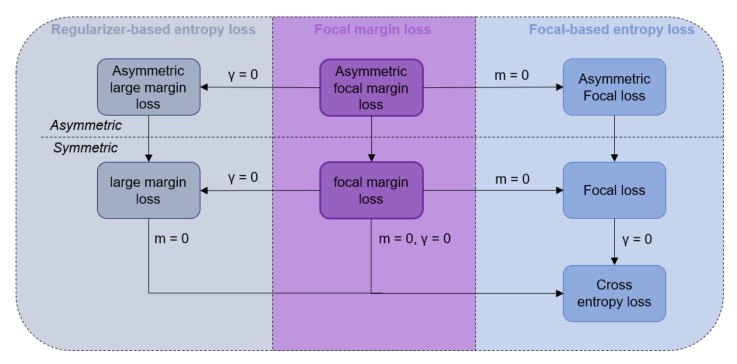} 
\caption{Our proposed focal margin loss unifies the regularizer-based entropy loss and the focal-based entropy loss. The arrows and associated parameters indicate the transitions between the formulas. The pink part in the middle represents the main contribution. The focal margin loss produces the focal loss when margin value ($m$) is zero, and produces the large margin loss when the $\gamma$ is zero. The focal margin loss yields the standard cross-entropy loss when both $m$ and $\gamma$ are set to zero.} 
\label{Fig.1} 
\end{figure}

\begin{table*}[htbp]
	\renewcommand{\arraystretch}{1.3}
        \caption{Experimental results for the DeepCrack Dataset with 237 test images, our proposal performs best in IoU compared to previous losses.}
	\label{tab1}
	\centering
	\begin{tabular}{cccccc}
	\hline
	& & DeepCrack-DB && \\
	\hline
	Loss functions &Parameters &IoU &F1 & Recall & Precision \\
	\hline
	$ L_{bce}$& - & 60.67&75.43&66.98&\textbf{87.01}\\
	$ L_{dice}$& - & 67.49&80.49&77.44&84.38\\
	$ L_{FL}$& $\hat{\gamma}=2.0$ & 56.38&71.99&63.14&84.74\\
	$ L_{aFL}$& $\hat{\gamma}=2.0$ & 61.94&76.41&73.41&80.38\\
	$ L_{Tversky}$& $\delta=0.7$ & 68.27&81.05&82.46&80.16\\
	$ L_{FT}$& $\delta=0.7, \gamma=0.75$ &68.47&81.19&83.71&79.26\\
	$ {L_{bcedice}}^{\mathrm{a}}$& - &68.33 & 81.10 &80.17 &82.59 \\
	$ {L_{HF}}^{\mathrm{b}}$& $\hat{\gamma}=2.0,\delta=0.7,\gamma=0.75$ & 69.32&81.80&82.82&81.30\\
	$ {L_{aUF}}^{\mathrm{c}}$& $\delta=0.6,\gamma=0.5$ & 68.91&81.51&80.55&82.94\\
	\hline
	$ L_{ours}{\mathrm{d}}$& $\hat{\gamma}=2.0,\delta=0.7,\gamma=0.75,m=0.5$ & \textbf{69.75{\color{green}(+0.43)}}&\textbf{82.09{\color{green}(+0.29)}}&83.57{\color{green}(+0.75)}&81.09\\
	$ L_{ours}$& $\hat{\gamma}=2.0,\delta=0.7,\gamma=0.75,m=1.0$ & 69.62{\color{green}(+0.30)} &82.00{\color{green}(+0.20)} &\textbf{83.71}{\color{green}(+0.89)} &80.77\\
	$ L_{ours}$& $\hat{\gamma}=2.0,\delta=0.7,\gamma=0.75,m=1.5$ & 69.38{\color{green}(+0.06)} & 81.83{\color{green}(+0.03)} & 82.98{\color{green}(+0.16)} &81.19\\
	\hline
	\end{tabular}
	\vspace{0.3cm}
\end{table*}

\begin{table*}[htbp]
	\renewcommand{\arraystretch}{1.3}
        \caption{Experimental results for the CRACK500 Dataset with 494 test images, our proposal performs the same as previous $L_{bcedice}$.}
	\label{tab2}
	\centering
	\begin{tabular}{cccccc}
	\hline
	& & CRACK500 && & \\
	\hline
	Loss functions &Parameters &IoU &F1 & Recall & Precision \\
	\hline       
	$ L_{bce}$& - & 54.10&70.17&64.99&\textbf{76.42}\\
	$ L_{dice}$& - & 59.34&74.44&75.24&73.78\\
	$ L_{FL}$& $\hat{\gamma}=2.0$ & 46.95&63.84&55.33&75.74\\
	$ L_{aFL}$& $\hat{\gamma}=2.0$ & 52.89&69.15&68.36&70.16\\
	$ L_{Tversky}$& $\delta=0.7$ & 57.47&72.95&82.45&65.55\\
	$ L_{FT}$& $\delta=0.7, \gamma=0.75$ &57.33&72.84&\textbf{82.67}&65.21\\
	$ {L_{bcedice}}^{\mathrm{a}}$& - & 60.65&75.47&76.37&74.73\\
	$ {L_{HF}}^{\mathrm{b}}$& $\hat{\gamma}=2.0,\delta=0.7,\gamma=0.75$ & 59.10&74.25&81.90&68.01\\
	$ {L_{aUF}}^{\mathrm{c}}$& $\delta=0.6,\gamma=0.5$ & 59.35&74.45&79.42&70.18\\
	\hline
	$ L_{ours}{\mathrm{d}}$& $\hat{\gamma}=0.0,\delta=0.5,\gamma=1.0,m=0.5$ & \textbf{60.66}{\color{green}(+0.01)} &\textbf{75.48}{\color{green}(+0.01)} &77.40{\color{green}(+1.03)} &73.79\\
	$ L_{ours}$& $\hat{\gamma}=0.0,\delta=0.5,\gamma=1.0,m=1.0$ & 60.54&75.38&77.66 &73.38\\
	$ L_{ours}$& $\hat{\gamma}=0.0,\delta=0.5,\gamma=1.0,m=1.5$ & 60.60& 75.43 & 77.86 &73.28\\
	\hline
	\end{tabular}
	\vspace{0.3cm}
\end{table*}

\begin{table*}[htbp]
	\renewcommand{\arraystretch}{1.3}
        \caption{Experimental results for the PanelCrack Dataset with 513 test images, our proposal performs best in IoU compared to previous losses.}
	\label{tab3}
	\centering
	\begin{tabular}{cccccc}
	\hline
	& & PanelCrack && & \\
	\hline
	Loss functions &Parameters &IoU &F1 & Recall & Precision \\
	\hline       
	$ L_{bce}$& - & -&-&-&-\\
	$ L_{dice}$& - & 23.22&37.55&57.26&28.43\\
	$ L_{FL}$& $\hat{\gamma}=2.0$ & 4.88&9.17&5.06&32.12\\
	$ L_{aFL}$& $\hat{\gamma}=2.0$ & 11.83&20.93&13.46&41.54\\
	$ L_{Tversky}$& $\delta=0.7$ & 33.51&50.00&59.10&43.63\\
	$ L_{FT}$& $\delta=0.7, \gamma=0.75$ &33.10&49.56&59.24&42.87\\
	$ {L_{bcedice}}^{\mathrm{a}}$& - & 32.09&48.45&49.58&\textbf{47.60}\\
	$ {L_{HF}}^{\mathrm{b}}$& $\hat{\gamma}=2.0,\delta=0.7,\gamma=0.75$ & 34.64&51.24&56.60&47.14\\
	$ {L_{aUF}}^{\mathrm{c}}$& $\delta=0.6,\gamma=0.5$ & 32.72&49.08&51.19&47.53\\
	\hline
	$ L_{ours}{\mathrm{d}}$& $\hat{\gamma}=2.0,\delta=0.7,\gamma=0.75,m=0.5$ & 34.94{\color{green}(+0.3)} &51.55{\color{green}(+0.31)} &56.87\color{green}(+0.27) &47.49\\
	$ L_{ours}$& $\hat{\gamma}=2.0,\delta=0.7,\gamma=0.75,m=1.0$ & 34.85{\color{green}(+0.21)}&51.49{\color{green}(+0.25)} &57.57\color{green}(+0.97) &46.95\\
	$ L_{ours}$& $\hat{\gamma}=2.0,\delta=0.7,\gamma=0.75,m=1.5$ & \textbf{35.08}{\color{green}(+0.44)} & \textbf{51.74}{\color{green}(+0.50)} & \textbf{57.61}{\color{green}(+1.01)} &47.32\\
	\hline
	\multicolumn{4}{l}{$^{\mathrm{a}}$The performance of vanilla loss used in \cite{OptimizedCrack} for crack segmentation.}\\
    \multicolumn{4}{l}{$^{\mathrm{b}}$A combination of asymmetric focal loss and focal tversky loss.}\\
    \multicolumn{4}{l}{$^{\mathrm{c}}$Asymmetric Unified Focal Loss with default parameters.}\\
    \multicolumn{4}{l}{$^{\mathrm{d}}$Focal Margin component in our loss produces asymmetric focal loss when $m=0$.}
	\end{tabular}
	\vspace{0.3cm}
\end{table*}

Observing the foreground and background terms of \eqref{eq.2} and \eqref{eq.3} respectively, the regularized foreground term of $L_{aM}$ can be combined with the weight attenuation background term of $L _{aF}$, leading to the following asymmetric focal margin loss:
\begin{equation}
    \mathop{L _{aFM}}\limits_{c=1}(P,T)  = - \frac{1}{N}T_{c}\cdot\log_{}{\hat{P_c}}  - \frac{1}{N}P_{c}^{\hat{\gamma}}\cdot (1-T_{c})\cdot{\log_{}{(1-P_{c})} }\label{eq.11}
\end{equation}
Here, N represents a number of samples, Pc and Tc represent prediction and ground truth of class c, respectively. $\hat{P_c}$ is the regularized prediction. The removed loss attenuation for foreground class pushes it away from the decision boundary, and the added margin regularizer further moves the decision boundary closer to the background, which makes it suitable to handle extremely imbalanced data such as crack data and preventing overfitting.
$ L _{aFM}$ degenerates to $L_{aM}$ for $\hat{\gamma}=0$. With $m=0$, \eqref{eq.11} yields $L_{aF}$. Moreover, setting both $\hat{\gamma}$ and $m$ to 0, $L _{aFM}$ becomes the standard binary cross-entropy loss. The symmetric focal margin loss is given as follows:
\begin{equation}
    \mathop{L _{FM}}(P,T)  = - \frac{1}{N}(1-P_{c})^{\hat{\gamma}}\cdot T_{c}\cdot \log_{}{\hat{P_c}}
\end{equation}

Therefore, the hybrid focal loss in \cite{Focus-Unet} is optimized by replacing the focal loss with our proposed focal margin loss. The final optimized hybrid focal margin loss is given as follows:
\begin{equation}
    L_{sHFM} = \lambda L_{FM} + (1-\lambda)L_{FT}
\end{equation}In the binary class segmentation, the non-rare term in $L_{aFT}$ can be discarded which simplifies to $L_{FT}$. In addition, the $\delta$ of focal loss component in \cite{Unified-Focal} is removed in our case since the contribution of rare foreground loss is small enough, thus the weight of rare foreground does not need to be attenuated while the background weight can still be reduced by the suppression parameter $\hat{\gamma}$. Moreover, Yeung indicates that $\lambda$ is partially redundant \cite{Unified-Focal} should be simplified as well. To this end, we simplify our optimized loss ($L_{ours}$) for experimental purposes as follows, and assume that the results showing an increase or decrease in performance also carry over to other derivations of dice and entropy-based losses:
\begin{equation}
    L_{ours} = L_{aFM} + L_{FT}
\end{equation}
A version of hybrid focal loss used in our experiments for comparison purpose is defined as follows:
\begin{equation}
    {L}_{HF} = L_{aF} + L_{FT}
\end{equation}

\section{Experiments}
\paragraph{Datasets} We use the open-sourced DeepCrack Dataset (DeepCrack-DB) \cite{DeepCrack-Liu}. It contains 537 images (300 for training, and 237 for testing) with sizes of 554*384 pixels. In order to reduce the deformation effect caused by scaling and speed up the experiments, we simply cropped each image to 384*384, and adjusted the training and test images to 96*96 pixels. Cracks account for 5.05\% of the resampled DeepCrack dataset. 

Another dataset for road cracks, CRACK500 \cite{RoadCrack, Featurepyramid}, was used in our experiments. Due to the average image size being 1509*2512 pixels, we first picked 250 images from CRACK500, then cropped images to 512*512, and finally resized the crack images to 128*128 pixels. All images containing no cracks were removed. The experimental CRACK500 dataset (7.3\% is crack) contains 1481 training images (75\%) and 494 testing images (25\%).

Our PanelCrack dataset consists of industrial panel cracks is also added, which contains 3.15\% crack images. We created panel cracks partly by hand tapping and partly by collecting real panel crack data from the factory and scanning it with a camera. A total number of 2051 images were resampled into 128*128 pixels, with 1538 images (75\%) for training and 513 images (25\%) for testing.

\begin{figure}[htbp]
\setlength{\belowcaptionskip}{0.1cm}   
\centering 
\includegraphics[width=0.45\textwidth]{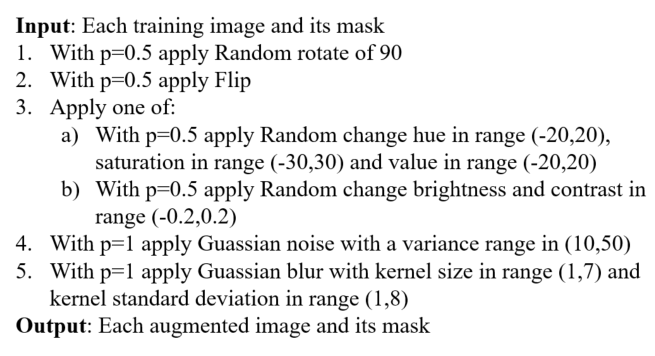} 
\caption{Detailed augmentation policy for our training experiments. p repressents the probability with which each augmentation step is applied.} 
\label{Fig.2} 
\end{figure}

\paragraph{Augmentations} A standard augmentation policy shows in Fig.~\ref{Fig.2} was applied to all experiments, alleviating the over-fitting in the beginning.

\paragraph{Loss Function}
To investigate the effect of different margins, we use the default $\delta=0.7$ suggested by \cite{Focaltversky} and the default $\hat{\gamma}=2.0$ suggested by \cite{Overfitting} for the asymmetric focal loss component. In our opinion, a $\hat{\gamma}$ value larger than 1.0 can suppress the background efficiently due to the activation of logits are between (0,1). The exponent $\gamma$ for $L_{FT}$ is 0.75 as suggested by \cite{Focaltversky}, which enhances the rare class. Table~\ref{tab1},~\ref{tab2} and ~\ref{tab3} show the performance of $ L_{ours}$ with different margin values on DeepCrack-DB, CRACK500 and our datasets, respectively. Moreover, the baseline loss ($L_{bcedice}$) and other previous loss functions, e.g., $L_{FL}$, $L_{aFL}$, $L_{Tversky}$, $L_{FT}$, $L_{HF}$ and $L_{aUF}$ are also presented for a comparison purpose.

\paragraph{Analysis}
We have tried different margin values from 0 to 2 for each dataset and found that 0.5, 1.0 and 1.5 obtained outstanding IoU on the three experimental datasets, respectively. Hence, 0.5, 1.0 and 1.5 are used as a set of hyperparameters in formal experiments. All losses are evaluated by using the recent Crack Segmentation Architecture \cite{OptimizedCrack} with Unet basis. Each model is trained 10 times (each time for 100 epochs) and all these results are averaged to a mean value. As shown in Table~\ref{tab1} for margin values larger than 0, further improvements of segmentation results are achieved. Specifically, when $m=0.5$, the recall of DeepCrack-DB increases 0.75, which also leads to the increase of IoU and F1. However, $L_{aUF}$ with the default parameters suggested by \cite{Unified-Focal} performs worse than the experimental $L_{HF}$.

For CRACK500 dataset, $L_{bcedice}$ performs best due to crack accounted for a relatively high proportion in CRACK500 compared to other datasets in our experiments. Other parameters are therefore simplified except $m$ to $L_{bcedice}$. Note that with $m=0$, hybrid focal margin loss produces hybrid focal loss. In the experiments, $m$ does not significantly improve the segmentation and the results are almost equivalent with or without $m$.
\begin{figure}[!htb]
\setlength{\belowcaptionskip}{0.1cm}   
\centering 
\includegraphics[width=0.5\textwidth]{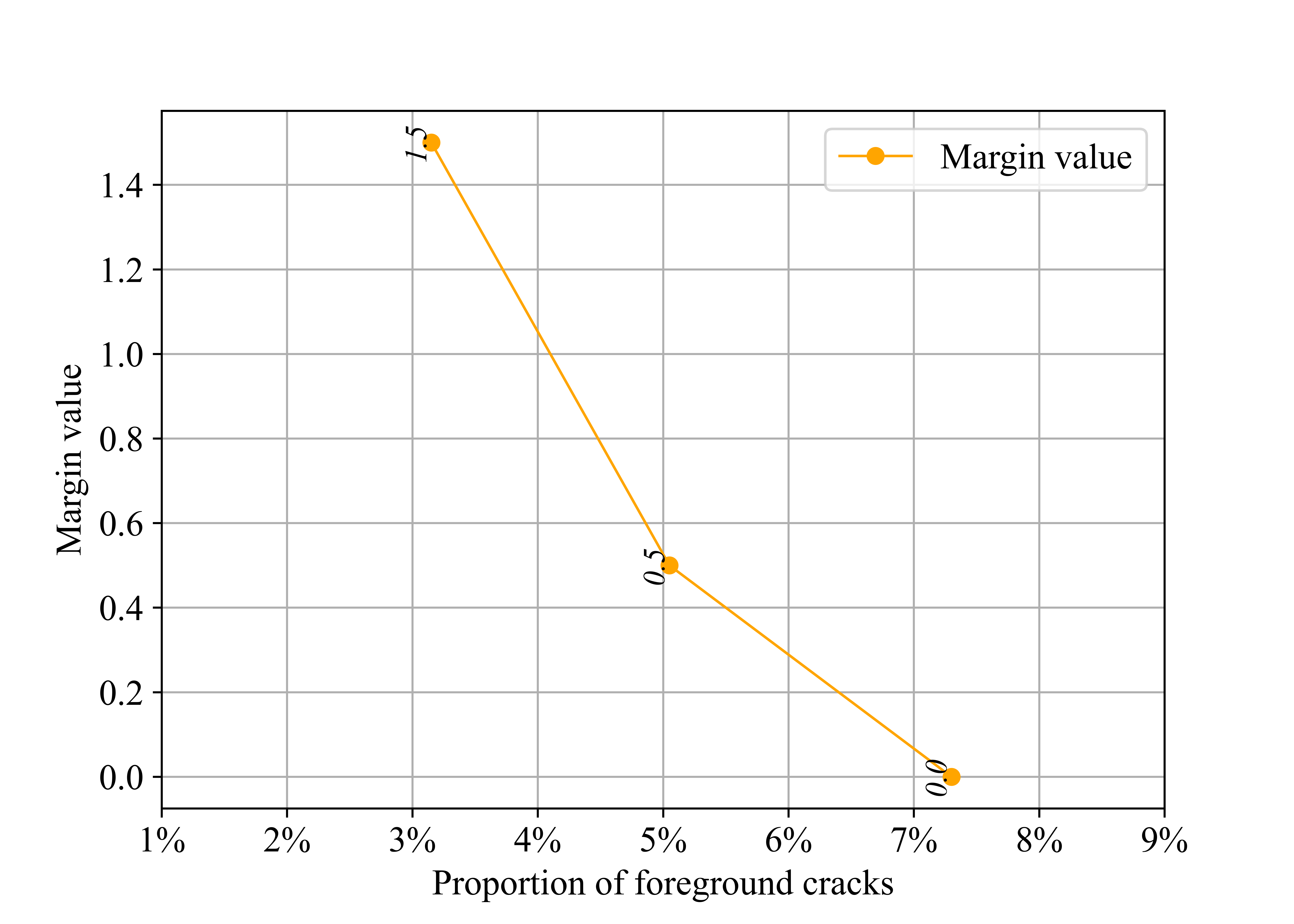} 
\caption{The choice of margin value depends on the degree of class imbalance. A higher margin value than 1.5 is recommended when the proportion of foreground is less than 3\%. A lower margin value than 0.5 is also recommended when the proportion of foreground is larger than 7\%.} 
\label{Fig.3} 
\end{figure}
Table~\ref{tab3} shows the model performance on our dataset. $L_{bce}$ does not work due to extreme class imbalance. With $m=1.5$, our proposal improves Recall by 1.01, F1 by 0.5 and IoU by 0.44, respectively. According to our experiments, margin value in the range of 0.5 to 1.5 is recommended for class imbalance segmentation. Moreover, a large margin value is recommended when the class imbalance is more severe and the training data is less. In general, the dice-based loss performs better than the entropy-based loss, especially when the data is extremely imbalanced. That leads to the less contribution of entropy component to the whole compound loss. The previous focal-based entropy loss alleviates this problem to some extent, while our proposal further improves the entropy component of the compound loss and thus boosts the overall segmentation performance. Fig.~\ref{Fig.4} illustrates the outperformance of our proposal compared to previous losses.

In addition, the original pretrained EfficientUnet-B7 and TransUnet-R50-ViT-B\_16 have also been tested on DeepCrack-DB using our loss function and previous compound losses for comparison. Table~\ref{tab4} shows that our proposal is also suitable for other networks such as EfficientUNet and TransUNet.
\begin{table}[!htb]
	\renewcommand{\arraystretch}{1.3}
        \caption{Our proposal still improves the IoU on the DeepCrack-DB that uses the recent EfficientUnet and TransUnet architectures. Each result on this table is the mean value of 10 times attempts.}
	\label{tab4}
	\centering
\resizebox{8.5cm}{1.0cm}{
\begin{tabular}{|l|l|l|l|l|}
\hline
Models  & $L_{bcedice}$ & $L_{HF}$ & $L_{aUF}$ & $L_{ours}$ \\ \hline
CrackSeg \cite{OptimizedCrack}     & 68.33 &    69.32  &     68.91   &  \textbf{69.75}  \\ \hline
EfficientUNet-B7 \cite{Eff-UNet} & 67.28 & 67.43 &     67.39   &\textbf{67.73}    \\ \hline
TransUNet-R50-ViT-B\_16 \cite{Transunet} &64.21& 68.02&64.24&  \textbf{68.84}    \\ \hline
\end{tabular}}
\end{table}
\begin{figure*}[htbp]
\centering 
\includegraphics[width=0.8\textwidth]{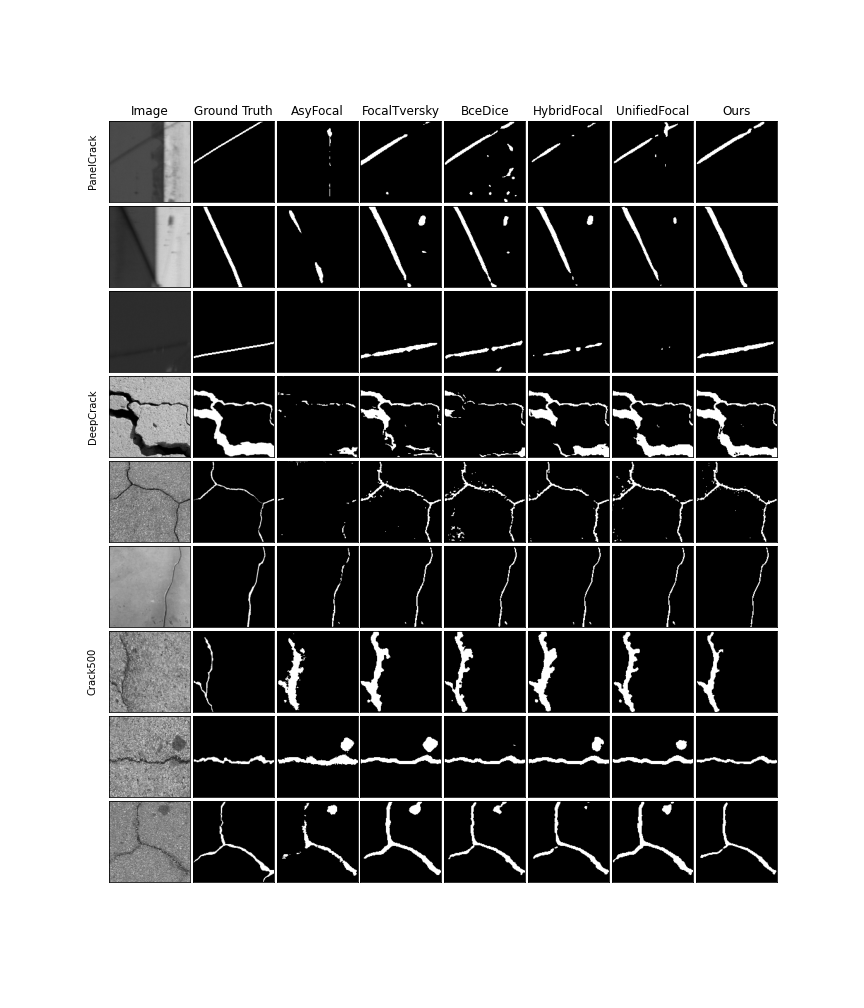} 
\caption{Comparison of AsyFocal, FocalTversky, BceDice, HybridFocal, UnifiedFocal and Our proposal, The results of Hybrid Focal Margin loss are closest to ground truth in most cases.} 
\label{Fig.4} 
\end{figure*}

\section{Conclusion}
In this work, we investigate previous loss functions for class-imbalanced data. We reveal that entropy-based loss can be further split into regularizer-based entropy loss and focal-based entropy loss and propose an optimized hybrid focal margin loss to optimize the previous losses. In a complex background, the Focal Margin component can not only address class imbalance, but also prevent overfitting. Our experiments demonstrate that margin values in the range of 0.5 to 1.5 are recommended for image segmentation with different degrees of class imbalance. Our proposed method outperforms the baseline BceDice and the HybridFocal in IoU scores and presents balanced precision-recall scores. We believe that the Focal Margin component and its modifications and combination with other losses can handle various segmentation tasks flexibly.


\end{document}